\documentclass[11pt,a4paper]{article}
\usepackage[hyperref]{emnlp2020}
\usepackage{times}
\pdfoutput=1
\usepackage{latexsym}

\usepackage{graphicx}
\usepackage{amsfonts}
\usepackage{microtype}
\usepackage{array}
\usepackage{booktabs}
\usepackage{algorithm}
\usepackage{color}
\usepackage{bm}
\usepackage{algorithmic}

\aclfinalcopy 

\title{RDSGAN: Rank-based Distant Supervision Relation Extraction with Generative Adversarial Framework}

\author{Guoqing Luo\\
    School of Computer Science \\
    Wuhan University\\
    {\tt guoqingluo@whu.edu.cn}\\\And
    Jiaxin Pan\\
    School of Computer Science \\
    Wuhan University\\
    {\tt pjx\_1997@whu.edu.cn}\\\And
    Min Peng\thanks{*Min Peng is the Corresponding Author} \\
    School of Computer Science \\
    Wuhan University\\
    {\tt pengm@whu.edu.cn}\\}

\date{}

\begin{document}
\maketitle
\begin{abstract}
\vspace{-0.15cm}
Distant supervision has been widely used for relation extraction but suffers from noise labeling problem. Neural network models are proposed to denoise with attention mechanism but cannot eliminate noisy data due to its non-zero weights. Hard decision is proposed to remove wrongly-labeled instances from the positive set though causes loss of useful information contained in removed instances. In this paper, we propose a novel generative neural framework named \textbf{RDSGAN} (Rank-based Distant Supervision GAN) which automatically generates valid instances for distant supervision relation extraction. Our framework combines soft attention and hard decision to learn the distribution of true positive instances via adversarial training and selects valid instances conforming to the distribution via rank-based distant supervision, which addresses the false positive problem. Experimental results show the superiority of our framework over strong baselines.

\end{abstract}
\section{Introduction}
\vspace{-0.1cm}
Relation extraction is fundamental for constructing large scale knowledge bases, which aims to extract the relations between entity pairs. One popular way to handle this task is distant supervision \cite{mintz2009distant} which automatically generates numerous labeled data via aligning text with the existing knowledge bases. However, generated training data contains numerous noisy samples due to the strong assumption. To tackle this issue, most recent state-of-the-art methods perform neural networks \cite{du2018multi,li2019gan,beltagy2019combining} on denoising operation with distant supervision. Various attention mechanisms \cite{lin2016neural,han2018hierarchical,gao2019hybrid} are proposed for calculating precise attention weights over instances, but soft attention mechanism usually assigns non-zero weights to noisy instances, which does not eliminate noisy data. \citet{Qin2018DSGAN,qin2018robust,ma2019easy} argue that wrongly-labeled instances must be treated with hard decision by removing false positive instances from the positive set, though hard decision may cause loss of useful information contained in removed instances. In order to keep as much useful information and reduce as much noise as possible, combining both soft attention and hard decision to learn the distribution of true positive instances is a better choice.

In this paper, we propose a novel generative neural framework Rank-based Distant Supervision GAN (named RDSGAN). Firstly, we train the framework to learn the distribution of true positive instances excluding false positive instances via adversarial training. Secondly, we rank all the instances in a sentence bag and select instances conforming to the distribution of true positive instances with the method of rank-based distant supervision, which optimizes the framework to generate a
clean and valid instance in each sentence bag and addresses the false positive problem. Finally, the framework can automatically generate massive valid instances\footnote{Valid instances include true positive and true negative instances} and thus provide a clean dataset for distant supervision relation extraction.

Our contributions are summarized as follows:

(1) We propose a novel generative neural framework which learns the distribution of true positive instances and automatically generates massive valid instances to provide a clean dataset for distant supervision relation extraction.

(2) We propose the method of rank-based distant supervision to address the false positive problem.

\begin{figure*}[ht]
	\begin{center}
\includegraphics[scale=0.28]{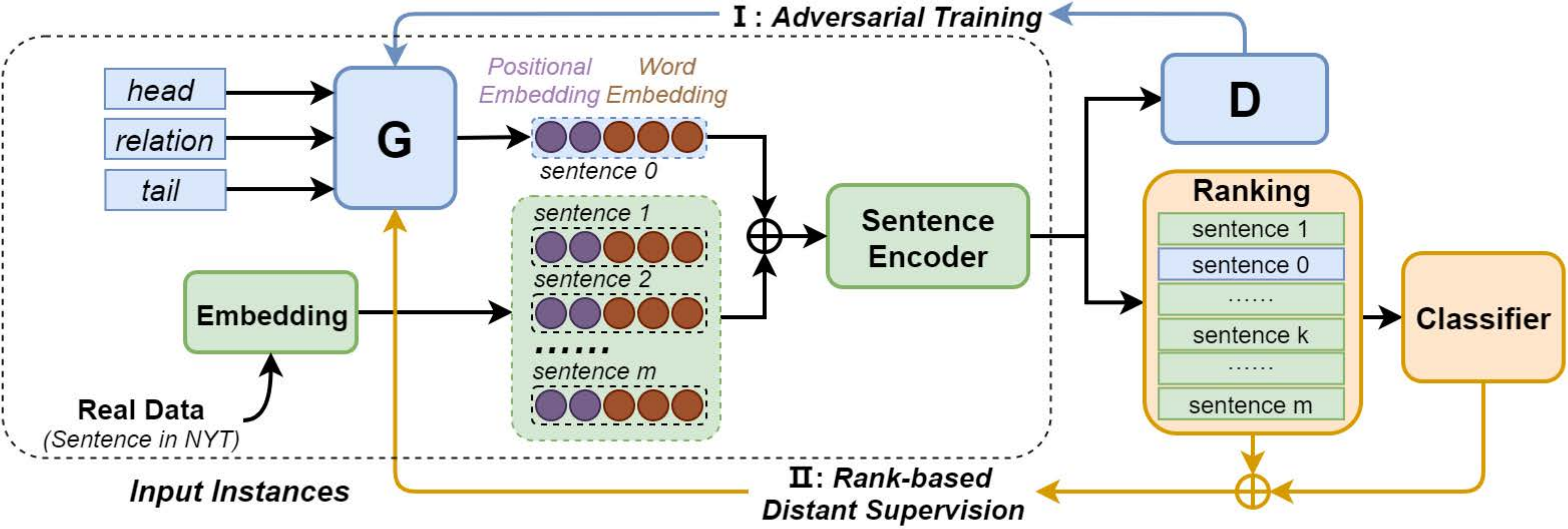}
	\end{center}
	\vspace{-0.3cm}
	\caption{Overview of RDSGAN. Input instances are the concatenation (denoted by $\boldsymbol{\oplus}$) of encoded embeddings including: a) \textit{sentence 0} from the generator (denoted by \textbf{G}) with the triplet $(head,relation,tail)$, and b) \textit{sentence 1} to $m$ from instances in NYT dataset. Firstly, input instances are fed into the discriminator (denoted by \textbf{D}) for adversarial training. Secondly, we fix D and rank instances in the ranking module (denoted by \textbf{Ranking}) and also perform relation classification for rank-based distant supervision. Please see Section \ref{method} for more details.}
	\label{architecture}
\end{figure*}
\vspace{-0.1cm}

\section{Methodology} \label{method}
In this section, we present the procedure of our framework, details of adversarial training and rank-based distant supervision as follows.
\subsection{Framework}
As illustrated in Figure \ref{architecture}, input instances are the concatenation of encoded embeddings of \textit{sentence 0} to $m$, we initialize the discriminator (D) and the generator (G) with random weights $\theta_d$ and $\theta_g$. In the first phase, input instances are fed to train D to learn the distribution of true positive instances, then G is trained to generate instances more similar to real ones. In the second phase, we fix D and use ranking module to rank mixed instances, then we select instances conforming to the distribution based on selective attention \cite{lin2016neural}, which produces the bag representation for relation classification. Rank loss $\mathcal{L}_1$ and relation classification loss $\mathcal{L}_2$ are added (denoted by $\color{orange} \boldsymbol{\oplus}$) with weights to optimize G to generate a valid instance in one bag for building up a clean dataset for distant supervision. The complete training procedure of the framework is shown in Algorithm \ref{training}.
\begin{algorithm}[htb]
\caption{Algorithm of RDSGAN}
\label{training}
\begin{algorithmic}[1] 
\REQUIRE $D,G,\mathcal{L}_1,\mathcal{L}_2$, $s_D,s_G$ and $s_R$ are  iterator numbers of each module
\ENSURE
\FOR{numbers of training iterations}
\FOR{$s_D$ steps}
\STATE Fix $G$, update $D$ by:
\STATE \small{$\nabla_{\theta_d}[\frac{1}{M_i}\sum_{i=1}^{M_i} (\log D(\mathbf{x}_i)+\log(1- D(\mathbf{x}_i))]$}
\ENDFOR
\FOR{$s_G$ steps}
\STATE Fix $D$, update $G$ by:
\STATE $\nabla_{\theta_g}\frac{1}{M_i}\sum^{M_i}_{i=1}\log(1- D(\mathbf{x}_i))$
\ENDFOR
\FOR{$s_R$ steps}
\STATE Fix $D$, update $G$ based on equation \ref{sum}:
\STATE $\nabla_{\theta_g}(\lambda_1\mathcal{L}_1+\lambda_2\mathcal{L}_2)$
\ENDFOR
\ENDFOR
\end{algorithmic}
\end{algorithm}
\subsection{Adversarial Training}

\subsubsection{Generator}
The target of the generator is to generate a vector sequence representing a clean and valid instance which conforms to the distribution of true positive data. As shown in Figure \ref{architecture}, The decoder-based generator is fed into a triplet $(h,r,t)$ and outputs a valid vector sequence. Hence, given the triplet of $(h,r,t)$, we first map $h$ and $t$ into vectors via their word embeddings and map $r$ via a relation matrix $\mathbf{A} \in \mathbb{R}^{N_r \times d_s}$, i.e. $\mathbf{e}_r = \mathbf{Ar}$, where $N_r$ is the number of all relation classes, and $d_s$ is the dimension of sentence embedding, $\mathbf{r}$ is the query vector associated with relation $r$. The input of the generator is the sum of the three vectors:
\vspace{-0.1cm}
\begin{equation}
\mathbf{z} = \mathbf{e}_h + \mathbf{W}_g\mathbf{e}_r + \mathbf{e}_t
\end{equation}

In detail, we utilize Bidirectional-GRU (BiGRU) for the decoder and place dropouts on the hidden states of BiGRU. The generation process can be formulated as:
\begin{equation}
\begin{array}{l}
\mathbf{h}_{i+1} = BiGRU(\mathbf{h}_i)
\end{array}
\end{equation}
where $\mathbf{h}_i \in \mathbb{R}^{d}$ is the hidden vector of the BiGRU and $\mathbf{h}_0 = \mathbf{z}$. The generation process goes on until it reaches the aligned sentence length $L$. After the generation, we obtain a sentence bag $\mathbf{X} = \{ \mathbf{x}_0,\mathbf{x}_1,\cdots,\mathbf{x}_m\}$ shown in Figure \ref{architecture}, then we feed the sentence bag into the discriminator.

\subsubsection{Discriminator}
The discriminator is designed to learn the distribution of the true positive data, for each instance in a sentence bag, the discriminator calculates its probability of coming from the real data as follows:
\vspace{-0.1cm}
\begin{equation}
\mathcal{L}_D(\mathbf{x}_i,\theta_d) =\log D(\mathbf{x}_i) + \log(1- D(\mathbf{x}_i))
\end{equation}
where $i=0,1,\cdots,m$ and $m$ is the number of instances in a bag. Hence, as for instances $\mathbf{x}$ in the $j$-th bag $M_j$ in the training data, the discrimination loss $\mathcal{L}_D$ can be formulated as:
\begin{equation}
\mathcal{L}_{D(M_j)} = \sum_{\mathbf{x_i}\in M_j} (\log D(\mathbf{x}_i) + \log(1- D(\mathbf{x}_i))
\end{equation}
\subsection{Rank-based Distant Supervision}

As shown in Figure \ref{architecture}, Ranking and Classifier perform rank-based distant supervision. Given a bag $M$ containing $m$ instances related to entity pair $(h, t)$, the representation of $M$ and the conditional probability of $(h,t)$ expressing relation $r$ are respectively calculated as:
\begin{equation}\label{eq:array}
\hspace{-0.2em}\mathbf{q} = \sum_{i=1}^m \alpha_i \mathbf{x}_i,
\hspace{0.5em}
p(r|M;\Theta) = \frac{\exp(o_r)}{\sum_{i=1}^{N_r} \exp(o_i)}
\end{equation}
where $\mathbf{q}$ is the representation of $M$, $\alpha_i$ is the attention weight for each sentence $\mathbf{x}_i$. $N_r$ is the total number of relation classes. $\Theta$ represents all the parameters, and $o_r$ is the score for relation $r$,:
\begin{equation}
\mathbf{o} = \mathbf{W}_r \mathbf{q} + \mathbf{b}_2
\end{equation}
where $\mathbf{W}_r$ is weight matrix and $\mathbf{b}_2$ is a bias vector.

We further define the loss function for rank-based distant supervision as the sum of rank loss $\mathcal{L}_1$ and relation classification loss $\mathcal{L}_2$ with their respective weights $\lambda_1$,$\lambda_2>0$:
\vspace{-0.1cm}
\begin{equation}
\mathcal{L} = \lambda_1\mathcal{L}_1 + \lambda_2\mathcal{L}_2 \label{sum}
\end{equation}
\textbf{Rank Loss:} In the ranking module, for all the instances in one bag, an instance containing less or no noise has higher attention weights and thus ranks higher. Hence, we attempt to make the generated instance rank in top-$k$ ($k$ is a hyperparameter), and rank loss of the generated instance $\mathcal{L}_{rank}^G$ in a bag is calculated as follows:
\vspace{-0.1cm}
\begin{equation}
    \mathcal{L}_{rank}^G=\frac{\exp(e_i)}{\sum_{i=1}^k \exp(e_i)}
\end{equation}
where $e_i$ is referred to as a query-based function which scores how well the input instance $\mathbf{x}_i$ and the predicting relation $\mathbf{r}$ matches. The rank loss $\mathcal{L}_1$ can be calculated as the average of the rank loss of each bag, where $m$ is the number of instances in a sentence bag:
\vspace{-0.2cm}
\begin{equation}
\mathcal{L}_1 = \frac{1}{m}\sum_{i=1}^{N_b}\mathcal{L}_{rank}^G
\end{equation}
\textbf{Relation Classification Loss:} We define the loss of relation classification $\mathcal{L}_2$ using cross-entropy:
\vspace{-0.1cm}
\begin{equation}
\mathcal{L}_2 = \sum_{i=1}^{N_b}\log p(r_i|M_i;\Theta)
\end{equation}

\section{Experiments}

\subsection{Experiment Setup}
We conduct experiments on Riedel dataset \cite{riedel2010modeling}, which aligns Freebase relations with the New York Times (NYT) corpus. The dataset contains 53 relations including no relation ``NA''. There are 522,611 sentences linked to 281,270 entity pairs for training and 172,448 sentences linked to 96,678 entity pairs for testing.

In our experiments, we adopt stochastic gradient descent (SGD) as optimization strategy. We select the word dimension as $50$, position dimension as $10$, kernel size as $3$, the number of feature maps or filters as $230$, batch size as $160$, aligned sentence length $L$ as $120$, and the dropout probability as $0.5$. We also set the learning rate of generator and discriminator as $1e-5$ and $1e-4$ respectively.

Following previous works, we evaluate our framework on the held-out evaluation. We adopt Precision@N (P@N), area under curve (AUC) and aggregated Precision-Recall (PR) curves as evaluation metrics to illustrate the performance of our proposed framework.

\subsection{Performance Evaluation of RDSGAN}
We adopt following baselines for distant supervised relation extraction.\\
• \textbf{Mintz} \cite{mintz2009distant}, \textbf{MultiR} \cite{hoffmann2011knowledge} and \textbf{MIML} \cite{surdeanu2012multi}: Non-neural models based on handcrafted features.\\
• \textbf{CNN+ATT} and \textbf{PCNN+ATT} \cite{lin2016neural}: Robust CNN-based models reducing noisy data based on selective attention mechanism.\\
• \textbf{DSGAN+ATT} \cite{Qin2018DSGAN}: A robust model using GAN to recognize true positive data.\\
• \textbf{PDCNN+TATT} \cite{peng2019dilated}: A dilated CNN-based model with soft entity type constraints.

The overall performance of our method compared with aforementioned baselines for distant supervised relation extraction is shown in Table \ref{PR curves}. We can see that our method achieves much better results on P@N (100, 200, 300) metrics, and improves the AUC value by 8.98\% and 7.69\% compared to DSGAN+ATT and PDCNN+ATT respectively. The huge improvement comes from rank-based distant supervision which reduces much false positive data for relation extraction.
\begin{table}[th]
\begin{center}
\resizebox{0.5\textwidth}{!}{
\begin{tabular}{lccccc}
\hline
P@N& 100& 200& 300& Mean & AUC \\
\hline
CNN+ATT& 76.2& 68.6& 59.8& 68.2& 0.33\\
PCNN+ATT&76.2 &73.1 &67.4& 72.2& 0.35\\
DSGAN+ATT&78.0& 75.5& 72.3& 75.3& 0.35\\
PDCNN+TATT&83.2&81.1&76.4&80.2&0.36\\\hline
\textbf{RDSGAN+ATT}&\textbf{88.9}&\textbf{85.3}&\textbf{81.1}&\textbf{85.1}&\textbf{0.39}\\
\hline
\end{tabular}
}
\caption{Overall performance at P@Ns(\%) and AUC values of different models on the NYT dataset}
\label{PR curves}
\end{center}
\end{table}
\vspace{-0.3cm}
\begin{figure}[th]
	\begin{center}
		\resizebox{0.5\textwidth}{!}{
		\includegraphics[scale=0.95]{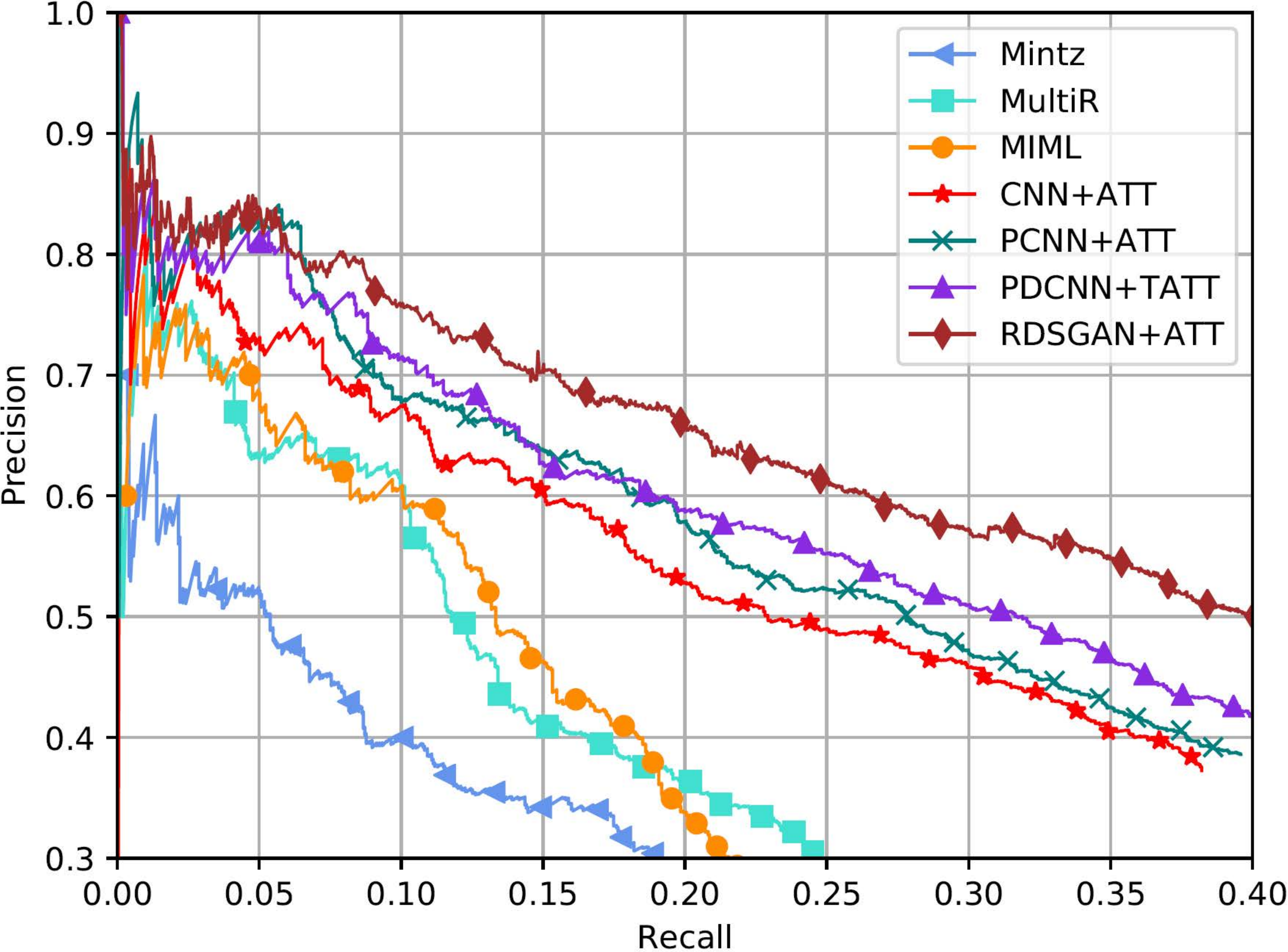}
		}
	\end{center}
 	\vspace{-0.2cm}  
	\caption{\label{figure4}Comparison of PR curves between our proposed model and baselines on the NYT dataset}
\end{figure}

We also plot PR curves between different models shown in Figure \ref{figure4} with recall number smaller than 0.4. From the overall result, we can see that: (1) All the non-neural baselines perform poorly as their features used by them are mostly derived from NLP tools, which can be erroneous. (2) CNN+ATT and PCNN+ATT improve the performance because they utilize sentence-level selective attention to reduce noise in the bag of entity pair. (3) PDCNN+TATT further enhances the performance as it incorporates soft entity type constraints to improve attention mechanism. (4) Our method RDSGAN+ATT achieves the best precision over the entire range of recall on the NYT dataset. As the recall rate increases, the precision rate of RDSGAN+ATT decreases more slowly than other models and outperforms PDCNN+ATT by 6\% on average. It shows that our proposed framework can consistently generate valid instances to promote the performance for distant supervision relation extraction.

\section{Related Work}
\textbf{Generative Adversarial Training:} Recent studies have proposed several GAN-based methods utilizing gradient information in adversarial training to generate instances for relation extraction. \citet{Qin2018DSGAN} proposes DSGAN to recognize true positive instances from noisy dataset via reinforcement learning \cite{yu2017seqgan}. \citet{li2019gan} uses GAN-driven semi-distant supervision approach to construct accurate instances and avoid wrong negative labeling. \citet{zhao2019auxiliary} proposes an auxiliary classifier in the discriminator to generate high-quality training data for relation classifiers. Unlike previous models focusing on discrimination, we focus on generating valid instances to provide a clean dataset for relation extraction.

\textbf{Neural Relation Extraction}: In recent years, neural network models have shown superior performance on denoising operation over relation extraction. \citet{zhang2018attention} explores the attention-based capsule networks in a multi-instance multi-label learning (MIML) framework. \citet{bai2019structured} employs minimally structured learning to predict instance-level relation mentions. \citet{beltagy2019combining} utilizes joint training on distant supervision to identify noisy sentences. Most recently, BERT \cite{devlin2018bert} and its variants \cite{shi2019simple,soares2019matching,papanikolaou2019deep} have been proposed to leverage attention mechanism and transformer to learn word contextual relations. Unlike previous approaches, we utilizes rank-based distant supervision which combines both soft attention and hard decision to reduce noise.

\section{Conclusion}
In this paper, we propose RDSGAN, a novel generative neural framework which learns the distribution of true positive instances and automatically generates massive valid instances to provide a clean dataset for distant supervision relation extraction. We propose the method of rank-based distant supervision to address the false positive problem. Experimental results on the NYT dataset shows the superiority of our framework over strong baselines.

\bibliography{emnlp2020}

\begin{thebibliography}{22}
\expandafter\ifx\csname natexlab\endcsname\relax\def\natexlab#1{#1}\fi

\bibitem[{Bai and Ritter(2019)}]{bai2019structured}
Fan Bai and Alan Ritter. 2019.
\newblock Structured minimally supervised learning for neural relation
  extraction.
\newblock \emph{arXiv preprint arXiv:1904.00118}.

\bibitem[{Beltagy et~al.(2019)Beltagy, Lo, and Ammar}]{beltagy2019combining}
Iz~Beltagy, Kyle Lo, and Waleed Ammar. 2019.
\newblock Combining distant and direct supervision for neural relation
  extraction.
\newblock In \emph{Proceedings of the 2019 Conference of the North American
  Chapter of the Association for Computational Linguistics: Human Language
  Technologies, Volume 1 (Long and Short Papers)}, pages 1858--1867.

\bibitem[{Devlin et~al.(2018)Devlin, Chang, Lee, and
  Toutanova}]{devlin2018bert}
Jacob Devlin, Ming-Wei Chang, Kenton Lee, and Kristina Toutanova. 2018.
\newblock Bert: Pre-training of deep bidirectional transformers for language
  understanding.
\newblock \emph{arXiv preprint arXiv:1810.04805}.

\bibitem[{Du et~al.(2018)Du, Han, Way, and Wan}]{du2018multi}
Jinhua Du, Jingguang Han, Andy Way, and Dadong Wan. 2018.
\newblock Multi-level structured self-attentions for distantly supervised
  relation extraction.
\newblock \emph{arXiv preprint arXiv:1809.00699}.

\bibitem[{Gao et~al.(2019)Gao, Han, Liu, and Sun}]{gao2019hybrid}
Tianyu Gao, Xu~Han, Zhiyuan Liu, and Maosong Sun. 2019.
\newblock Hybrid attention-based prototypical networks for noisy few-shot
  relation classification.

\bibitem[{Han et~al.(2018)Han, Yu, Liu, Sun, and Li}]{han2018hierarchical}
Xu~Han, Pengfei Yu, Zhiyuan Liu, Maosong Sun, and Peng Li. 2018.
\newblock Hierarchical relation extraction with coarse-to-fine grained
  attention.
\newblock In \emph{Proceedings of the 2018 Conference on Empirical Methods in
  Natural Language Processing}, pages 2236--2245.

\bibitem[{Hoffmann et~al.(2011)Hoffmann, Zhang, Ling, Zettlemoyer, and
  Weld}]{hoffmann2011knowledge}
Raphael Hoffmann, Congle Zhang, Xiao Ling, Luke Zettlemoyer, and Daniel~S Weld.
  2011.
\newblock Knowledge-based weak supervision for information extraction of
  overlapping relations.
\newblock In \emph{Proceedings of the 49th Annual Meeting of the Association
  for Computational Linguistics: Human Language Technologies-Volume 1}, pages
  541--550. Association for Computational Linguistics.

\bibitem[{Li et~al.(2019)Li, Zhang, Jia, and Zhao}]{li2019gan}
Pengshuai Li, Xinsong Zhang, Weijia Jia, and Hai Zhao. 2019.
\newblock Gan driven semi-distant supervision for relation extraction.
\newblock In \emph{Proceedings of the 2019 Conference of the North American
  Chapter of the Association for Computational Linguistics: Human Language
  Technologies, Volume 1 (Long and Short Papers)}, pages 3026--3035.

\bibitem[{Lin et~al.(2016)Lin, Shen, Liu, Luan, and Sun}]{lin2016neural}
Yankai Lin, Shiqi Shen, Zhiyuan Liu, Huanbo Luan, and Maosong Sun. 2016.
\newblock Neural relation extraction with selective attention over instances.
\newblock In \emph{Proceedings of the 54th Annual Meeting of the Association
  for Computational Linguistics (Volume 1: Long Papers)}, volume~1, pages
  2124--2133.

\bibitem[{Ma et~al.(2019)Ma, Wang, Feng, and Huai}]{ma2019easy}
Shuai Ma, Gang Wang, Yansong Feng, and Jinpeng Huai. 2019.
\newblock Easy first relation extraction with information redundancy.
\newblock In \emph{Proceedings of the 2019 Conference on Empirical Methods in
  Natural Language Processing and the 9th International Joint Conference on
  Natural Language Processing (EMNLP-IJCNLP)}, pages 3842--3852.

\bibitem[{Mintz et~al.(2009)Mintz, Bills, Snow, and
  Jurafsky}]{mintz2009distant}
Mike Mintz, Steven Bills, Rion Snow, and Dan Jurafsky. 2009.
\newblock Distant supervision for relation extraction without labeled data.
\newblock In \emph{Proceedings of the Joint Conference of the 47th Annual
  Meeting of the ACL and the 4th International Joint Conference on Natural
  Language Processing of the AFNLP: Volume 2-Volume 2}, pages 1003--1011.
  Association for Computational Linguistics.

\bibitem[{Papanikolaou et~al.(2019)Papanikolaou, Roberts, and
  Pierleoni}]{papanikolaou2019deep}
Yannis Papanikolaou, Ian Roberts, and Andrea Pierleoni. 2019.
\newblock Deep bidirectional transformers for relation extraction without
  supervision.
\newblock \emph{EMNLP-IJCNLP 2019}, page~67.

\bibitem[{Peng et~al.(2019)Peng, Hu, Tian, Wang, Wang, and
  Wang}]{peng2019dilated}
Min Peng, Weilong Hu, Gang Tian, Bin Wang, Hua Wang, and Gang Wang. 2019.
\newblock Dilated convolutional networks incorporating soft entity type
  constraints for distant supervised relation extraction.
\newblock In \emph{2019 International Joint Conference on Neural Networks
  (IJCNN)}, pages 1--7. IEEE.

\bibitem[{Qin et~al.(2018{\natexlab{a}})Qin, Xu, and Wang}]{Qin2018DSGAN}
Pengda Qin, Weiran Xu, and William~Yang Wang. 2018{\natexlab{a}}.
\newblock Dsgan: Generative adversarial training for distant supervision
  relation extraction.

\bibitem[{Qin et~al.(2018{\natexlab{b}})Qin, Xu, and Wang}]{qin2018robust}
Pengda Qin, Weiran Xu, and William~Yang Wang. 2018{\natexlab{b}}.
\newblock Robust distant supervision relation extraction via deep reinforcement
  learning.
\newblock \emph{arXiv preprint arXiv:1805.09927}.

\bibitem[{Riedel et~al.(2010)Riedel, Yao, and McCallum}]{riedel2010modeling}
Sebastian Riedel, Limin Yao, and Andrew McCallum. 2010.
\newblock Modeling relations and their mentions without labeled text.
\newblock In \emph{Joint European Conference on Machine Learning and Knowledge
  Discovery in Databases}, pages 148--163. Springer.

\bibitem[{Shi and Lin(2019)}]{shi2019simple}
Peng Shi and Jimmy Lin. 2019.
\newblock Simple bert models for relation extraction and semantic role
  labeling.
\newblock \emph{arXiv preprint arXiv:1904.05255}.

\bibitem[{Soares et~al.(2019)Soares, FitzGerald, Ling, and
  Kwiatkowski}]{soares2019matching}
Livio~Baldini Soares, Nicholas FitzGerald, Jeffrey Ling, and Tom Kwiatkowski.
  2019.
\newblock Matching the blanks: Distributional similarity for relation learning.
\newblock \emph{arXiv preprint arXiv:1906.03158}.

\bibitem[{Surdeanu et~al.(2012)Surdeanu, Tibshirani, Nallapati, and
  Manning}]{surdeanu2012multi}
Mihai Surdeanu, Julie Tibshirani, Ramesh Nallapati, and Christopher~D Manning.
  2012.
\newblock Multi-instance multi-label learning for relation extraction.
\newblock In \emph{Proceedings of the 2012 joint conference on empirical
  methods in natural language processing and computational natural language
  learning}, pages 455--465. Association for Computational Linguistics.

\bibitem[{Yu et~al.(2017)Yu, Zhang, Wang, and Yu}]{yu2017seqgan}
Lantao Yu, Weinan Zhang, Jun Wang, and Yong Yu. 2017.
\newblock Seqgan: Sequence generative adversarial nets with policy gradient.
\newblock In \emph{Thirty-First AAAI Conference on Artificial Intelligence}.

\bibitem[{Zhang et~al.(2018)Zhang, Deng, Sun, Chen, Zhang, and
  Chen}]{zhang2018attention}
Ningyu Zhang, Shumin Deng, Zhanlin Sun, Xi~Chen, Wei Zhang, and Huajun Chen.
  2018.
\newblock Attention-based capsule networks with dynamic routing for relation
  extraction.
\newblock \emph{arXiv preprint arXiv:1812.11321}.

\bibitem[{Zhao(2019)}]{zhao2019auxiliary}
Yun Zhao. 2019.
\newblock An auxiliary classifier generative adversarial framework for relation
  extraction.
\newblock \emph{arXiv preprint arXiv:1909.05370}.

\end{thebibliography}
\bibliographystyle{acl_natbib}

\end{document}